\documentclass{article}

\PassOptionsToPackage{numbers, sort&compress}{natbib}  

 \usepackage[main, final]{neurips_2026}

\usepackage[utf8]{inputenc} 
\usepackage[T1]{fontenc}    
\usepackage{hyperref}       
\usepackage{url}            
\usepackage{booktabs}       
\usepackage{amsfonts}       
\usepackage{nicefrac}       
\usepackage{microtype}      
\usepackage{xcolor}         

\usepackage{amsmath}    
\usepackage{amssymb}    
\usepackage{adjustbox}  
\usepackage{multirow}   
\usepackage{capt-of}    
\usepackage{float}
\usepackage{wrapfig}    
\usepackage{tabularx}
\usepackage{array}

\title{ReVMap: Vectorized Global Mapping via Connectivity-Aware Local Map Fusion}

\author{%
  Hongjae Shin\thanks{Equal contribution.} \quad
  Jungho Kim\footnotemark[1] \quad
  Donghyuk Kwak \quad
  Myeongjun Kim \\
  \textbf{Seunghoon Yu} \quad
  \textbf{Jiyong Oh} \quad
  \textbf{Jun Won Choi\thanks{Corresponding author.}} \\[0.6em]
  Seoul National University \\
}

\begin{document}

\maketitle

\vspace{-0.3in}
\begin{figure}[h] 
    \centering
    \includegraphics[width=0.93\linewidth]{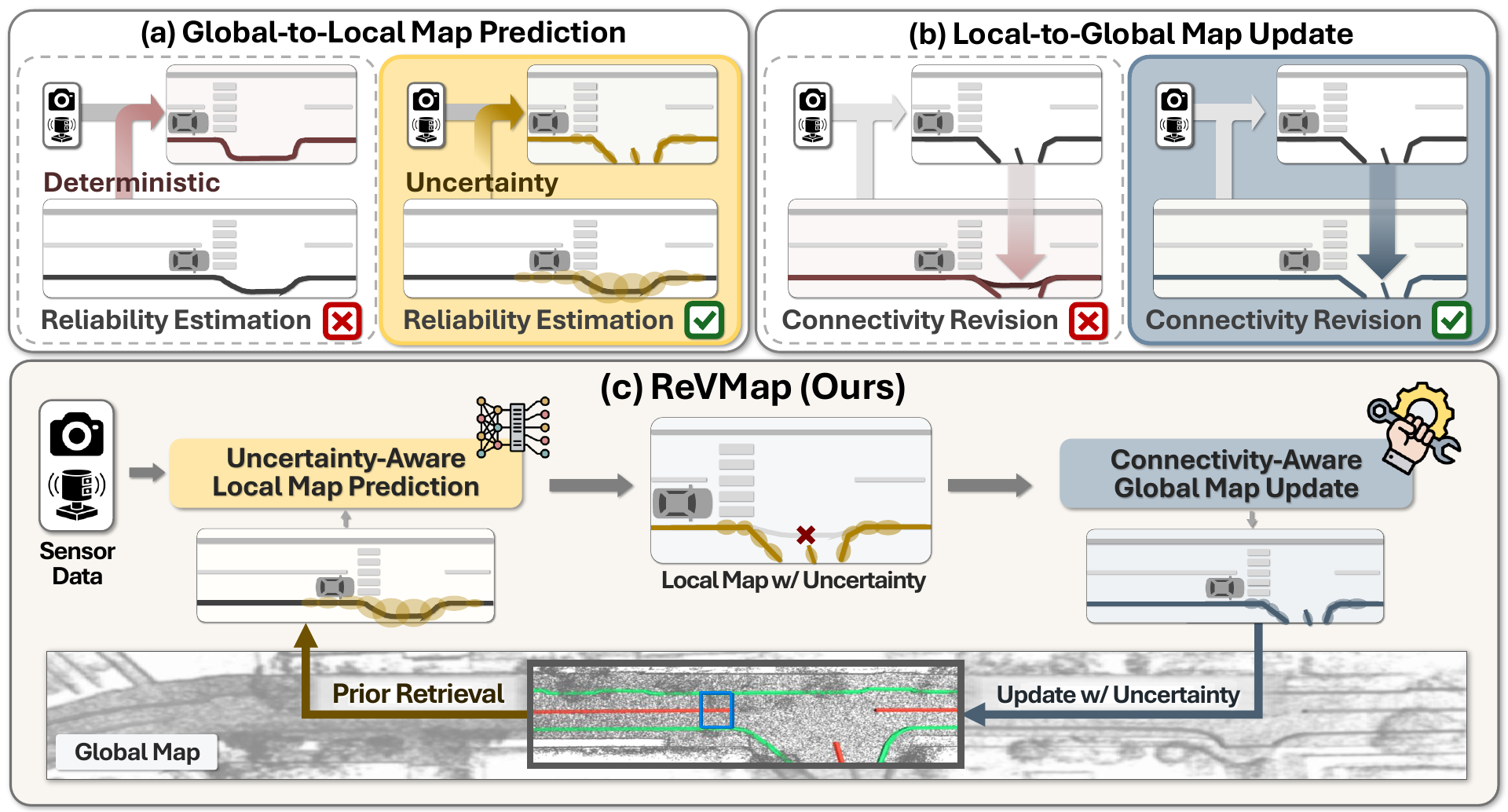}
    \caption{
    \textbf{Comparison of Map Update Paradigms and the ReVMap Framework.}
(a) Reliability-agnostic prior reuse leads to over-trusting unreliable accumulated elements, whereas uncertainty-aware prediction modulates their influence.
\textcolor{black}{(b) Heuristic matching can lead to incorrect map connections, whereas connectivity-guided revision enables more robust integration of local predictions into the global map.}
\textcolor{black}{(c) ReVMap iteratively constructs and refines the global map through uncertainty-aware local prediction and connectivity-guided global map revision.}
    }
    \label{fig:intro1} 
\end{figure}

\begin{abstract}

Large-scale annotation of vectorized high-definition (HD) maps is labor-intensive and costly, motivating local-to-global auto-labeling approaches that incrementally construct global maps from local vectorized predictions. However, existing approaches often overlook prediction uncertainty and rely on heuristic map integration, potentially propagating errors and introducing topological inconsistencies. To address these limitations, we propose ReVMap, a reliable local-to-global framework that iteratively refines local map predictions and global map construction through uncertainty-guided prior utilization and connectivity-guided global map revision. Specifically, ReVMap leverages point-wise localization uncertainty to incorporate global map priors into local map prediction, improving robustness against inaccurate prior information. It then predicts local-to-global connectivity and selectively updates the global map through uncertainty-aware fusion based on the reliability of local predictions. By iteratively coupling local prediction and global revision, ReVMap progressively improves both local map accuracy and global map consistency. Extensive experiments on the nuScenes and Argoverse 2 benchmarks demonstrate that ReVMap achieves state-of-the-art performance in both local map prediction and global map construction.
\end{abstract}

\section{Introduction}

HD maps are essential for safe autonomous driving, providing high-precision representations of lane boundaries, road edges, and road-element topology. Structured geometric and topological context from HD maps is widely used in downstream motion forecasting~\cite{lanegcn, rpred} and trajectory planning~\cite{uniad, vad, safedrive}. A typical HD map construction pipeline consists of three stages: 1) global scene reconstruction from collected sensor data, 2) vectorized annotation of road elements, and 3) quality verification before deployment. 
Within this pipeline, the vectorized annotation stage is particularly labor-intensive, as accurate labels require precise map-element geometry and consistent topology across road structures.
Recent HD map auto-labeling pipelines~\cite{vma, cama, camav2, thma} reduce manual labeling effort by producing vectorized HD map elements using learning-based map construction methods~\cite{maptr, maptrv2, mask2map, pivotnet, bemapnet, mapqr}, requiring human annotators only for minor corrections and quality verification.

For HD map auto-labeling in global coordinates, direct vector map generation over large areas is impractical due to high computational costs and the difficulty of optimizing long-range geometric and topological consistency. 
Previous works~\cite{globalmapnet, dumapnet, vma, cama, camav2, mvmap} therefore adopt a local-to-global formulation, decomposing the task into local vector map generation and global map integration. This decomposition improves scalability by focusing each prediction on a local observation range, thereby reducing computational and memory demands while preserving fine-grained map geometry. 
During global map integration, newly predicted local elements either replace outdated global elements corresponding to the same map structures or are added as new elements in regions not yet covered by the global map.

Another line of research~\cite{nmp, hrmapnet, presight, globalmapnet} explores the use of global map priors to improve local map prediction. Unlike offboard auto-labeling, where the global map primarily serves as the final supervisory target, these methods leverage the accumulated global map as prior information to guide subsequent local map estimation. Such priors provide valuable geometric and topological information that may be difficult to infer from current sensor observations alone, particularly in scenarios involving limited perception range, occlusions, or ambiguous road structures.

\textcolor{black}{However, these accumulated map priors may also contain errors from earlier local predictions. Existing methods~\cite{globalmapnet,vma,presight,hrmapnet} that leverage these priors often treat accumulated map elements as deterministic without explicitly accounting for their reliability. Inaccurate global map elements may thus be over-trusted, and their errors may propagate into subsequent local predictions (see Figure~\ref{fig:intro1}(a)).}

\textcolor{black}{A related challenge arises when integrating newly predicted local elements into the accumulated global map. Existing approaches~\cite{vma,globalmapnet} rely on heuristic rules, such as distance-based matching, to associate local predictions with existing global elements, which can lead to incorrect associations in the presence of prediction errors. They then update only the overlapping portions of matched global elements, leaving the remaining geometry unchanged. Consequently, incorrect associations may introduce topological inconsistencies, while partial updates may leave existing structural errors unresolved (see Figure~\ref{fig:intro1}(b)).}


\textcolor{black}{To address these challenges, we introduce ReVMap, a local-to-global framework for reliable vectorized global map construction, which can be used for HD map auto-label generation. As illustrated in Figure~\ref{fig:intro1} (c), ReVMap iteratively couples uncertainty-guided prior utilization with connectivity-guided global map revision. The framework comprises two core components: Uncertainty-aware Local Map Generation (ULG) and Global Map Revision Network (GMR).}

\textcolor{black}{ULG leverages the accumulated global map as a reliability-aware prior to guide local map prediction. It also estimates point-wise localization uncertainty for newly predicted local map elements, which is propagated to the global map through GMR. This enables the accumulated global map to maintain uncertainty-aware representations for subsequent local predictions.}


\textcolor{black}{GMR updates the accumulated global map using local predictions from ULG. For local-to-global association, GMR predicts connectivity between current local instances and surrounding global segments outside the local prediction region, rather than relying on heuristic matching rules. The predicted connectivity guides map revision, while uncertainty-based filtering and fusion reduce the influence of unreliable local predictions on the updated geometry. The updated global map and its uncertainty are then reused by ULG in subsequent iterations, establishing a continuous feedback loop between local map prediction and global map refinement.}


We evaluate ReVMap on large-scale HD map construction benchmarks, including nuScenes and Argoverse2. 
ReVMap achieves substantial performance gains over previous state-of-the-art methods on both benchmarks. 
On nuScenes, ReVMap establishes a new state of the art in both global map construction and local prediction, achieving 36.1\% mGAP, 43.4\% miECM, and 58.0\% local mAP.
On Argoverse2, ReVMap achieves 62.1\% mGAP, 62.5\% miECM, and 77.0\% local mAP, establishing new state-of-the-art performance across both global map construction and local prediction.

\begin{itemize}
    \item \textcolor{black}{We propose ReVMap, a local-to-global mapping framework that constructs and progressively refines a global map through uncertainty-guided prior utilization and connectivity-guided map revision.}

    \item We introduce ULG, which predicts vectorized local map elements and localization uncertainty to construct reliability-aware global map priors for subsequent local prediction.

    \item We present GMR that refines the global map through learned local-to-global connectivity estimation and uncertainty-aware map fusion.

    
    \item  ReVMap achieves state-of-the-art performance in both local map prediction and global map construction on the nuScenes and Argoverse 2 benchmarks.
\end{itemize}

\section{Related Work}
\label{related}

\noindent\textbf{Online Local Map Construction.}
Online local map construction~\cite{maptr, mapqr, maptrv2, hdmapnet, vectormapnet, mask2map, pivotnet, bemapnet, mgmap,instagram} predicts vectorized map elements around the ego vehicle from onboard sensor observations. 
Query-based approaches improve local map construction with structured map-element representations. 
MapTR~\cite{maptr} introduces permutation-equivalent point sets, and MapTRv2~\cite{maptrv2} strengthens supervision and matching. 
Mask2Map~\cite{mask2map} further incorporates BEV masks for instance-level semantics and point-level geometry. 
Since local prediction from current observations remains sensitive to limited perception range, occlusion, and temporal inconsistency, recent work incorporates accumulated map context into local prediction. 
NMP~\cite{nmp} utilizes global BEV features as a neural prior, HRMapNet~\cite{hrmapnet}
builds historical rasterized maps from previous predictions, and GlobalMapNet~\cite{globalmapnet} constructs an online global vector map for subsequent local prediction.

\noindent\textbf{Vectorized Global Map Construction.}
Vectorized global map auto-annotation and large-scale map generation~\cite{globalmapnet, vma, dumapnet, icmapper, egcvmap, unimapgen, thma} aim to produce spatially consistent vector annotations over large regions, often by integrating local predictions into accumulated global maps. 
MapTracker~\cite{maptracker} and IC-Mapper~\cite{icmapper} propagate element identities through tracking-based association, but their single-sequence update logic lacks mechanisms for existing-map revision or multi-trip integration. 
GlobalMapNet~\cite{globalmapnet} and VMA~\cite{vma} instead accumulate local predictions through geometric association, where GlobalMapNet adopts Chamfer-distance-based one-to-one matching and VMA adds endpoint-distance constraints. 
These hand-crafted criteria limit flexible matching under branching or fragmented structures, and both methods update only overlapping portions of matched elements, leaving inconsistent non-overlapping geometry unchanged after accumulation. 
CAMA and CAMAv2~\cite{cama, camav2} improve the consistency and accuracy of vectorized map auto-annotation from multi-camera data, but their VMA-style global update process inherits similar rule-based association and overlap-only update limitations.

\noindent\textbf{Uncertainty Estimation in HD Map Construction.}
Semantic confidence quantifies the model's certainty in the predicted class of a map element but does not directly capture its localization accuracy.
Explicit assessment of localization reliability is particularly important under occlusion and ambiguous road structures, motivating uncertainty estimation for map geometry. Prior work~\cite{uncertaintyTrajectory} leverages point-wise uncertainty to reduce the influence of unreliable map geometry on downstream trajectory prediction, and UIGenMap~\cite{uigenmap} introduces uncertainty-guided structure injection to improve the generalization of local map prediction. However, these approaches do not propagate uncertainty estimates into accumulated global maps. MV-Map~\cite{mvmap} employs an uncertainty network to estimate the per-frame prediction reliability for map aggregation, while Kalman-filter-based HD map updating~\cite{incrementalfusing} incorporates uncertainty into the fusion of observations. However, these methods do not maintain point-wise localization uncertainty in an accumulated vectorized global map for reuse in subsequent local prediction.

\section{Method}
\label{method}

\begin{figure}[t] 
    \centering
    \includegraphics[width=\linewidth]{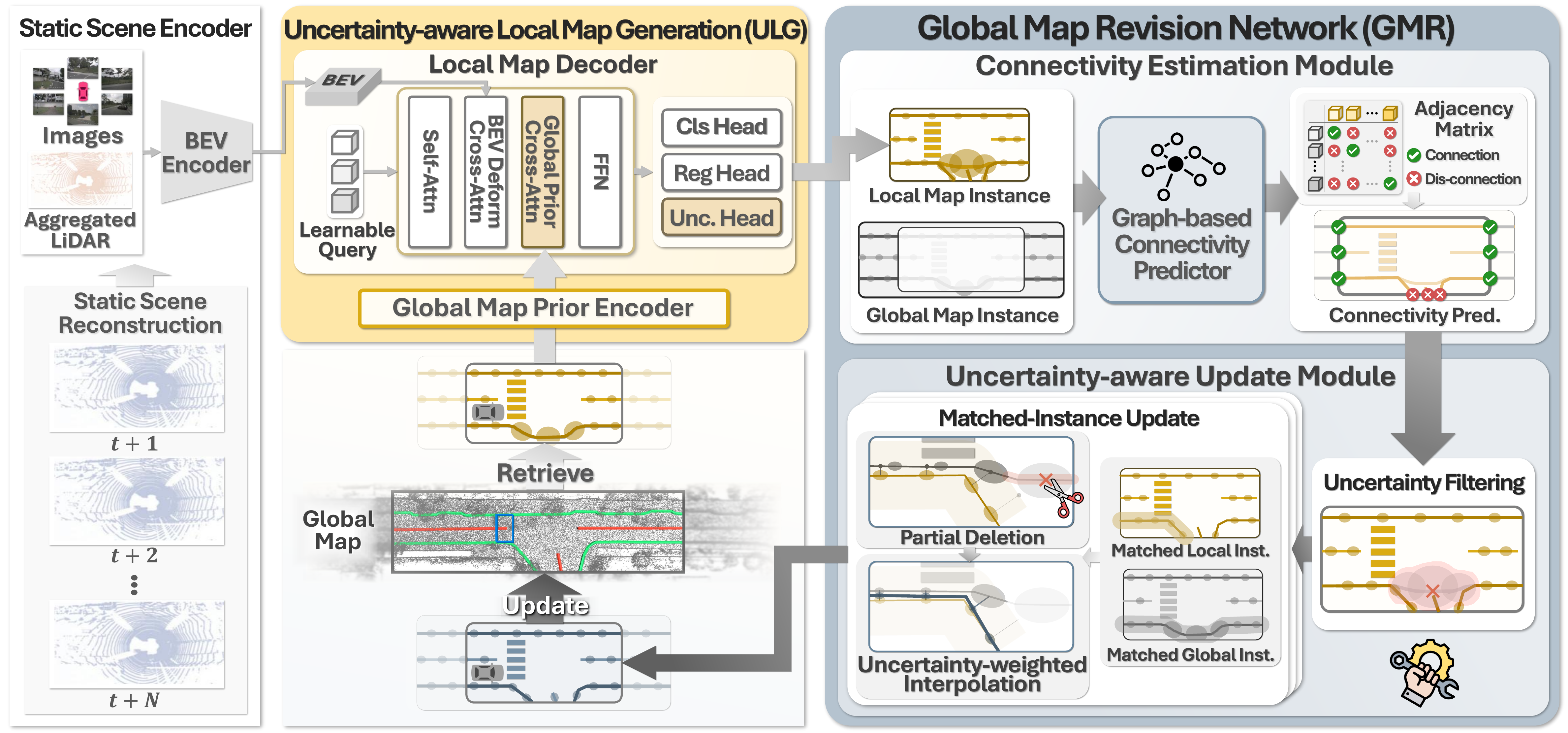}
    \caption{\textbf{Overall structure of ReVMap.} 
    ReVMap consists of three main components: Static Scene Encoder, ULG, and GMR. 
    The Static Scene Encoder generates a BEV feature map from multi-view images and,
    in the offboard setting, from a static 3D scene reconstructed from multi-timestep
    LiDAR sweeps.
    ULG retrieves global map priors and predicts local vectorized map instances with point-wise uncertainty. 
    GMR then estimates local-to-global connectivity and updates the accumulated global map via uncertainty-aware filtering and fusion.
    }
    \label{fig:main_method}
\end{figure}

\subsection{Overview}
Each vectorized map instance consists of a semantic class and an ordered sequence of 2D BEV points, where the classes are lane dividers, pedestrian crossings, and road boundaries.
Local instances use a fixed number of points, whereas accumulated global instances may contain a variable number of points after iterative merging and revision.
As illustrated in Figure~\ref{fig:main_method}, ReVMap consists of three components, Static Scene Encoder, ULG, and GMR, and operates iteratively over local prediction regions.
At each iteration, (1) the Static Scene Encoder encodes the reconstructed static scene and multi-camera images into BEV features. (2) ULG uses these features and the accumulated global map as a prior to predict local vectorized map instances with point-wise localization uncertainty. (3) GMR estimates local-to-global connectivity and revises the accumulated global map through uncertainty-aware updates. The updated global map and its uncertainty are reused as the prior for the next iteration.

\subsection{Static Scene Encoder}

\noindent\textbf{Static Scene Reconstruction.}
A single LiDAR sweep provides sparse and partially occluded observations of the road layout, which is insufficient for reliable vectorized global map construction. 
We therefore reconstruct a dense static scene by aggregating multi-timestep LiDAR sweeps in a common global coordinate frame. 
For each sweep, points belonging to dynamic objects are removed using per-frame GT 3D boxes before aggregation, preventing movable objects from being fused into the static map representation. 
The filtered sweeps are merged into a scene-level static point cloud, which provides stable geometric evidence for BEV feature extraction and vectorized map prediction.

\noindent\textbf{BEV Feature Extractor.}
The BEV feature extractor produces a unified static-scene BEV representation from the reconstructed static point cloud and multi-camera images. 
We encode the static point cloud with a SECOND encoder~\cite{second} and transform image features into BEV space using LSS~\cite{lss}. 
Since dynamic objects are removed only during LiDAR-based reconstruction, we exclude regions of dynamic objects during LSS pooling by identifying them using pretrained Mask2Former~\cite{mask2former}. 
Following BEVFusion~\cite{bevfusion}, the resulting camera BEV features are concatenated with the LiDAR BEV features and fused through a convolutional layer to obtain the final BEV feature map.

\subsection{Uncertainty-aware Local Map Generation (ULG)}
ULG consists of the Global Map Prior Encoder and Local Map Decoder. 
The Global Map Prior Encoder extracts reliability-aware prior embeddings from previously accumulated global map elements. The Local Map Decoder leverages global prior embeddings and the BEV feature map to predict local map instances with point-wise uncertainty.

\begin{figure}[t] 
    \centering
    \includegraphics[width=\linewidth]{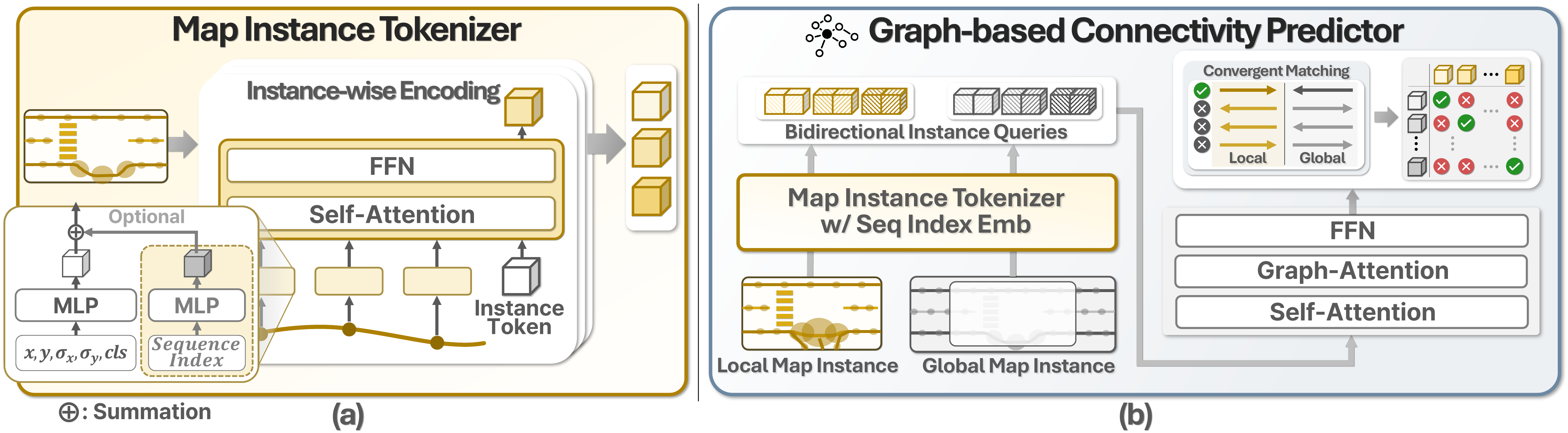}
    \vspace{-10pt}
    \caption{\textbf{Details of Map Instance Tokenizer and Graph-based Connectivity Predictor.} (a) The Map Instance Tokenizer aggregates per-point embeddings into a single instance token via self-attention. The sequence-index positional embedding is optional and applied only in the Graph-based Connectivity Predictor. (b) The Graph-based Connectivity Predictor first generates bidirectional instance queries from local and global instances using sequence-index positional embeddings, and then predicts their connectivity through self-attention, graph attention, and convergent matching.
    }
    \label{fig:method_detail}
\end{figure}

\noindent\textbf{Global Map Prior Encoder.}
The Global Map Prior Encoder tokenizes accumulated global map instances to provide global context to the Local Map Decoder. 
We apply the Map Instance Tokenizer illustrated in Figure~\ref{fig:method_detail} (a), which is inspired by the CLS-token-based aggregation mechanism~\cite{vit}. 
Accumulated global instances are represented as vector geometries with varying numbers of points, since they are formed by merging multiple local predictions. 
Before tokenization, each global instance is cropped to the local prediction region and uniformly resampled into $K$ points along its geometry, matching the fixed $K$-point format of local map instances. 
Each resampled point is first projected into a point embedding from its coordinate, class, and point-wise uncertainty using an MLP, without adding point-order positional embeddings.  
For each instance, a learnable instance token is concatenated to the point embeddings, and the combined tokens are processed by self-attention and FFN. 
The final instance token is used as the global prior embedding for the Local Map Decoder.

\noindent\textbf{Local Map Decoder with Uncertainty Estimation.}
The Local Map Decoder predicts the class, point coordinates, and point-wise localization uncertainty for each local map instance utilizing global prior embeddings. 
Following MapTRv2~\cite{maptrv2}, learnable map queries are decoded through multiple layers. 
In each decoder layer, the map queries exchange information through self-attention and aggregate local BEV features via deformable attention. 
The updated queries then interact with the global prior embeddings through cross-attention within the same layer, incorporating accumulated map context. 
Since the global prior embeddings encode accumulated uncertainty with geometry and class, this cross-attention enables the decoder to selectively leverage reliable priors when refining local predictions. 

The resulting prior-enhanced queries are fed into classification and regression heads. 
For localization, the regression head parameterizes each point as a Gaussian distribution, where the mean $\mathbf{p} \in \mathbb{R}^{2}$ represents the point coordinate and the standard deviation $\boldsymbol{\sigma} \in \mathbb{R}^{2}$ represents its localization uncertainty. 
The predicted uncertainty is propagated to the global map and later used for uncertainty-aware integration and prior reuse.

\subsection{Global Map Revision Network (GMR)}
\label{sec:grnet}

\textcolor{black}{GMR updates the accumulated global map within the current local prediction region while preserving geometry outside it. It first estimates connectivity between current local instances and surrounding global segments outside this region, and then updates the global geometry using point-wise uncertainty.}

\noindent\textbf{Connectivity Estimation Module.}
To prevent potentially erroneous geometry within the local prediction region from affecting connectivity estimation, we extract candidate segments from accumulated global instances within a fixed-margin band immediately outside the region. The Connectivity Estimation Module then predicts connections between these global segments and the current local instances.
As illustrated in Figure~\ref{fig:method_detail} (b), the Graph-based Connectivity Predictor estimates connectivity based on instance orientation and distance.
Since the same vector geometry can be traversed in two opposite directions, a single order-specific representation cannot capture all possible directed connections.
We therefore encode each instance in both its original and reversed point orders using the Map Instance Tokenizer with sequence-index embeddings, producing two direction-aware instance tokens for each instance.
Specifically, each point in a map instance is first projected from its coordinate, class, and point-wise uncertainty into a point embedding through an MLP. The sequence index is then embedded with another MLP and added to the point embeddings using the original and reversed point orders. The tokenizer outputs one directional instance token from each order, allowing the module to distinguish the two possible connection directions of each instance.

To explicitly model inter-instance distance, we update these instance tokens using self-attention and graph attention. 
Following~\cite{smart, hivt}, we construct the graph with edges connecting nearby instance pairs, and define the edge feature as the minimum pairwise point distance between two instances.
We predict a connectivity score for each directed local--global token pair and use these scores to determine local-to-global connections. 
For connectivity supervision, we adopt convergent matching, assigning a positive label only to the direction in which the two endpoints face each other. 
All unmatched pairs and non-converging directions are assigned negative labels.

\noindent\textbf{Uncertainty-aware Map Update Module.}
To prevent noisy geometry from being accumulated into the global map, the Uncertainty-aware Map Update first filters local instances using point-wise uncertainty. 
A local instance is discarded when its mean point-wise uncertainty from ULG is above an uncertainty threshold. 
For each remaining local instance, we restrict the update of its matched global instance to the local prediction region. 
Within this region, we delete global points whose nearest distance to the local instance is larger than a distance threshold, eliminating accumulated geometry that is inconsistent with the current local geometry.

The remaining global geometry is then fused with the local instance through point-wise uncertainty-weighted interpolation. 
Since interpolation is performed point-wise, we first establish correspondences between local and global points. 
For each local point, we construct a normal line passing through the point and perpendicular to the local tangent direction, and take its intersection with the remaining global geometry as the corresponding global point.
Since point-wise uncertainty is parameterized as the Gaussian standard deviation, interpolation weights are defined by inverse variance, $w = 1/(\sigma^2 + \epsilon)$, assigning larger weights to points with lower uncertainty. 
The corresponding local and global points are then interpolated with the normalized weights to update the global geometry.



\subsection{Training Loss}

The total loss $\mathcal{L}$ used to train ReVMap is given by
\begin{equation}
    \mathcal{L} = \mathcal{L}_{\mathrm{nll}} + \mathcal{L}_{\mathrm{cls}} + \mathcal{L}_{\mathrm{con}} + \mathcal{L}_{\mathrm{aux}},
\end{equation}
where $\mathcal{L}_{\mathrm{nll}}$ and $\mathcal{L}_{\mathrm{cls}}$ supervise the local map predictions of ULG, $\mathcal{L}_{\mathrm{con}}$ supervises the connectivity predictions of GMR, and $\mathcal{L}_{\mathrm{aux}}$ provides auxiliary depth and segmentation supervision following~\cite{maptrv2}.
The negative log-likelihood loss $\mathcal{L}_{\mathrm{nll}}$ supervises the point-wise uncertainty by modeling each predicted point as a bivariate Gaussian with mean $\hat{\mathbf{p}}_i$ and standard deviation $\boldsymbol{\sigma}_i$:
\begin{equation}
    \mathcal{L}_{\mathrm{nll}} = \frac{1}{N_p}\sum_{i=1}^{N_p}\sum_{k\in\{x,y\}} \left( \log \sigma_i^k + \frac{(p_i^k - \hat{p}_i^k)^2}{2(\sigma_i^k)^2}\right),
\end{equation}
where $\mathbf{p}_i$ is the ground-truth coordinate of the $i$-th point and $N_p$ is the number of points per instance. The classification loss $\mathcal{L}_{\mathrm{cls}}$ and the connectivity loss $\mathcal{L}_{\mathrm{con}}$ are defined as focal losses~\cite{focal} for class score prediction in ULG and connectivity prediction in GMR, respectively.

\begin{table*} [t!] 
\caption{Comparison with SOTA methods on the nuScenes new split validation set. “C” and “C+L” denote camera-only onboard setting and multi-modal offboard setting, respectively. The “-” indicates that the method does not build a scene-level global map and is included for comparison with local prediction.}
\vspace{3pt}
    \centering
  \begin{adjustbox}{width=0.95\textwidth}
    \begin{tabular}{l|c|c|ccc|c|ccc|c}
    \toprule[1.2pt]
    & & & \multicolumn{4}{c|}{GAP Metrics} & \multicolumn{4}{c}{iECM} \\
    \cmidrule(lr){4-7} \cmidrule(lr){8-11}
    Method & Modality & mAP & $div$ & $ped$ & $bou$ & $mean$ &  $div$ & $ped$ & $bou$ & $mean$\\ \toprule[0.7pt]
    {Mask2Map\cite{mask2map}} & C & 29.4 & - & - & - & - & - & - & - & - \\
    {HRMapNet\cite{hrmapnet}} & C & 37.1 & - & - & - & - & - & - & - & - \\
    {UniPrevPredMap\cite{uniprev}} & C & 39.7 & - & - & - & - & - & - & - & - \\
    \toprule[0.3pt]
    {StreamMapNet\cite{streammapnet}} & C & 32.9 & 13.4 & 15.5 & 12.8 & 13.9 & 32.8 & 16.3 & 40.8 & 30.0 \\
    {GlobalMapNet\cite{globalmapnet}} & C & 34.8 & 16.3 & 18.5 & 18.0 & 17.6 & 36.0 & 18.8 & 42.4 & 32.4 \\
    {MapTracker\cite{maptracker}}  & C & 40.3 & 18.1 & 36.1 & 19.4 & 24.5 & 38.7 & 27.4 & 45.2 & 37.1\\ 
    \toprule[0.7pt]
    {ReVMap (Ours)} & C & \textbf{45.6} & 18.1 & \textbf{37.9} & \textbf{29.6} & \textbf{28.5} & \textbf{42.3} & \textbf{27.7} & \textbf{50.9} & \textbf{40.3} \\ 
    \toprule[0.3pt]
    {ReVMap (Ours)} & C + L & \textbf{58.0} & \textbf{18.2} & \textbf{46.8} & \textbf{43.2} & \textbf{36.1} & \textbf{46.1} & \textbf{28.6} & \textbf{55.6} & \textbf{43.4}\\ 
    \bottomrule[1.2pt]
    \end{tabular}
  \end{adjustbox}
\label{tab:nusc_main}
\end{table*}

\begin{table*} [t!] 
\caption{Comparison with global map construction SOTA methods on Argoverse2 new split validation set. “C” and “C+L” denote camera-only onboard setting and multi-modal offboard setting, respectively.}
\centering
  \begin{adjustbox}{width=0.95\textwidth}
    \begin{tabular}{l|c|c|ccc|c|ccc|c}
    \toprule[1.2pt]
    & & & \multicolumn{4}{c|}{GAP Metrics} & \multicolumn{4}{c}{iECM} \\
    \cmidrule(lr){4-7} \cmidrule(lr){8-11}
    Method & Modality & mAP & $div$ & $ped$ & $bou$ & $mean$ &  $div$ & $ped$ & $bou$ & $mean$\\ \toprule[0.7pt]
    {GlobalMapNet\cite{globalmapnet}} & C  & 60.3 & 34.5 & 33.7 & 38.8 & 35.6 & 55.5 & 43.2 & 58.2 & 52.3 \\ 
    {MapTracker\cite{maptracker}}  & C  & 71.3 & 29.3 & 52.8 & 45.0 & 42.4  & 46.9 & 50.7 & 63.1 & 53.6\\ 
    \toprule[0.7pt]
    {ReVMap (Ours)} & C + L & \textbf{77.0} & \textbf{55.3} & \textbf{59.0} & \textbf{71.9} & \textbf{62.1} & \textbf{56.3} & \textbf{55.3} & \textbf{75.9} & \textbf{62.5} \\ 
    \bottomrule[1.2pt]
    \end{tabular}
  \end{adjustbox}
\label{tab:av2_qualitative}
\end{table*}


\section{Experiments}
\label{sec:experiment_results}
\subsection{Experiment Setup}
\label{sec:experiment}
\noindent\textbf{Datasets.}
We evaluate our method on nuScenes~\cite{nuscenes} and Argoverse2~\cite{argoverse2}, which contain 1,000 driving scenes each. nuScenes provides 20-second scenes at 2\,Hz, whereas Argoverse2 provides approximately 15-second scenes at 10\,Hz. To evaluate generalization to unseen areas, we follow the geographically non-overlapping new splits of StreamMapNet~\cite{streammapnet}.

\noindent\textbf{Evaluation Metrics.}
We evaluate local map prediction using \textit{mean Average Precision} (mAP), following previous works~\cite{maptrv2,streammapnet}. For global map construction, we adopt scene-level \textit{mean Global Average Precision} (mGAP)~\cite{globalmapnet}. We also report multi-trip mGAP, where overlapping predictions from different scenes are accumulated into an existing global map and evaluated as a unified map.
We further evaluate whether predictions cover each GT element without fragmentation by adapting the Entropy-based Connectivity Metric (ECM)~\cite{topoboundary,vma} to the instance level.
This penalizes fragmented predictions assigned to the same GT element, and the class-averaged score is reported as miECM. 
Details are provided in the Appendix.



\noindent\textbf{Implementation Details.}
For fair comparison, static point cloud reconstruction and dynamic feature masking are used only in the offboard camera--LiDAR setting, not in the onboard camera-only setting. 
Dynamic-object point clouds are removed using ground-truth 3D boxes, which can be replaced by any 3D object detector. 
We use ResNet-50 as the image backbone.
For the standard scene-level evaluation, we initialize each scene with a global map prior constructed from ReVMap's own local predictions across the scene.
ReVMap is trained in three stages: local-map-only training, connectivity-module training with the local model frozen, and end-to-end training with global priors, for 24, 1, and 6 epochs, respectively. More implementation details are provided in the Appendix.

\subsection{Performance Comparison}

Table~\ref{tab:nusc_main} compares ReVMap with recent online mapping and global map construction methods~\cite{mask2map,hrmapnet,uniprev,streammapnet,globalmapnet,maptracker} on the nuScenes new split validation set, showing state-of-the-art performance in the onboard camera-only setting and substantial gains with offboard camera--LiDAR inputs, reaching 58.0 mAP, 36.1 mGAP, and 43.4 miECM. Table~\ref{tab:av2_qualitative} shows that ReVMap also outperforms existing global map construction methods~\cite{globalmapnet,maptracker} on Argoverse2 across all metrics. 

Since the scene-level prior may contain elements observed only in later frames, Table~\ref{tab:causal} further compares camera-only methods with the same ResNet-50 backbone under a strictly causal protocol, where each scene starts from an empty map and the prior is built only from past frames.
ReVMap still outperforms MapTracker~\cite{maptracker} by 2.9\% mAP, 1.7\% mGAP, and 1.6\% miECM, showing that its gain does not rely on future observations.

Table~\ref{tab:nusc_cross_scene} evaluates multi-trip global map construction, a cross-scene setting that MapTracker~\cite{maptracker} cannot handle due to its single-scene identity propagation scheme. ReVMap outperforms GlobalMapNet~\cite{globalmapnet} in mAP, miECM, and mGAP, establishing state-of-the-art multi-trip global map construction.

\begin{table*}[t!]
\centering
\begin{minipage}[t]{0.48\textwidth}
\centering
\caption{Camera-only comparison on nuScenes with global map priors constructed using only past frames.}
\label{tab:causal}
\vspace{3pt}
\begin{adjustbox}{max width=\linewidth}
\begin{tabular}{l|ccc}
\toprule[1.2pt]
Method & mAP & mGAP & miECM \\
\midrule[0.7pt]
GlobalMapNet~\cite{globalmapnet} & 34.8 & 17.6 & 32.4 \\
MapTracker~\cite{maptracker} & 40.3 & 24.5 & 37.1 \\
ReVMap (Ours) & \textbf{43.2} & \textbf{26.2} & \textbf{38.7} \\
\bottomrule[1.2pt]
\end{tabular}
\end{adjustbox}
\end{minipage}
\hfill
\begin{minipage}[t]{0.48\textwidth}
\centering
\captionof{table}{Performance comparison of multi-trip global map
construction on nuScenes. The ``-'' indicates that the method cannot
merge maps across scenes. ``\(\ast\)'' denotes reproduced results.}
    \label{tab:nusc_cross_scene}
  \begin{adjustbox}{width=0.99\textwidth}
    \begin{tabular}{c|c|c|c}
    \toprule[1.2pt]
    Method & mAP & miECM & mGAP \\ 
    \toprule[0.7pt]
    GlobalMapNet\textsuperscript{*}~\cite{globalmapnet} & 34.8 & 25.5 &  13.7\\ 
    MapTracker~\cite{maptracker} & 40.3 & - & -  \\ 
    ReVMap (Ours) & \textbf{45.6} & \textbf{34.1} & \textbf{21.1}  \\ 
    \bottomrule[1.2pt]
    \end{tabular}
  \end{adjustbox}

\end{minipage}
\end{table*}

\begin{table*}[t!]
\vspace{-15pt}
\centering
\begin{minipage}{0.60\textwidth}
    \centering
    
    \captionof{table}{Ablation study of main components of ReVMap.}
    

\begin{adjustbox}{width=\textwidth}
\begin{tabular}{l|c|c|ccc|c}
\toprule[1.2pt]
& & & \multicolumn{4}{c}{GAP Metrics}\\
\cmidrule(lr){4-7}
Method & mAP & miECM & $div$ & $ped$ & $bou$ & $mean$ \\
\midrule
Baseline & 39.0 & 30.1 & 13.6 & 32.0 & 12.7 & 19.4 \\
+ Connectivity Estimation & 39.0 & 34.9 & 14.8 & 30.5 & 23.5 & 22.9 \\
+ Global Map Prior & 44.8 & 40.0 & 18.0 & 34.9 & 27.6 & 26.8 \\
+ Uncertainty Estimation & \textbf{45.6 } & 40.2 & 17.9 & 36.2 & 28.1 & 27.4 \\
+ Uncertainty Update & \textbf{45.6 } & \textbf{40.3} & \textbf{18.1} & \textbf{37.9} & \textbf{29.6} & \textbf{28.5} \\
\bottomrule[1.2pt]
\end{tabular}
\end{adjustbox}

    \label{nusc_main_ablation}
\end{minipage}
\hfill
\begin{minipage}{0.38\textwidth}
    \centering

    \captionof{table}{Effect of replacing the GT boxes used for dynamic-object removal in the C+L setting on nuScenes.}
    \label{tab:box_robust}
    \vspace{2pt}

    {\small
    \setlength{\tabcolsep}{5pt}
    \begin{tabular}{l|cc}
    \toprule[1.2pt]
    Boxes & mAP$\uparrow$ & mGAP$\uparrow$ \\
    \midrule[0.7pt]
    GT boxes & \textbf{58.0} & \textbf{36.1} \\
    Perturbed GT boxes & 57.6 & 35.6 \\
    No removal & 57.3 & 35.2 \\
    \bottomrule[1.2pt]
    \end{tabular}
    }
\end{minipage}
\end{table*}

\begin{table*}[t!]
\centering
\begin{minipage}[t]{0.46\textwidth}
    \centering
    \captionof{table}{Ablation study for evaluating the effect of Uncertainty Update. ``Filter'' denotes the Uncertainty Filtering and ``Interp'' refers to Uncertainty-weighted Interpolation.}

\begin{adjustbox}{width=\textwidth}
    \begin{tabular}{cc|c|cccc}
    \toprule[1.2pt]
    & & & \multicolumn{4}{c}{GAP Metrics} \\
    \cmidrule(lr){4-7}
    Filter & Interp & miECM & $div$ & $ped$ & $bou$ & $mean$ \\ 
    \midrule[0.7pt]
    & & 40.2 & 17.9 & 36.2 & 28.1 & 27.4 \\ 
    {\checkmark} &  & 40.2 & 18.0 & 37.3 & 29.0 & 28.1 \\ 
    & \checkmark & 40.3 & 17.8 & 37.1 & 28.9 & 27.9 \\ 
    \checkmark & \checkmark & \textbf{40.3} & \textbf{18.1} & \textbf{37.9} & \textbf{29.6} & \textbf{28.5} \\ 
    \bottomrule[1.2pt]
    \end{tabular}
\end{adjustbox}
    \label{ablation_uncertainty}
\end{minipage}
\hfill
\begin{minipage}[t]{0.52\textwidth}
    \centering
    \captionof{table}{Analysis of different local prediction and global construction method combinations.}
    \label{tab:cross_analysis}
  \begin{adjustbox}{width=\textwidth}
    \begin{tabular}{c|c|c|c|c}
    \toprule[1.2pt]
    Local Map & Global Map & mAP & miECM & mGAP \\ 
    \toprule[0.7pt]
    \multirow{2}{*}{GlobalMapNet} & GlobalMapNet & 34.8 & 32.4 & 17.6 \\ 
     & Ours & 34.8 & \textbf{35.2} \textcolor{red}{{\scriptsize(+2.8)}} & \textbf{19.4} \textcolor{red}{{\scriptsize(+1.8)}} \\ 
    \hline
    \multirow{2}{*}{MapTracker} & MapTracker & 40.3 & 37.1 & 24.5 \\ 
     & Ours & 40.3 & \textbf{37.8} \textcolor{red}{{\scriptsize(+0.7)}} & \textbf{26.5} \textcolor{red}{{\scriptsize(+2.0)}}    \\ 
    \hline
    \multirow{3}{*}{Ours} & GlobalMapNet & 45.6 & 38.2 & 23.5 \\ 
     & VMA & 45.6 & 39.3 & 25.7 \\ 
     & Ours & 45.6 & \textbf{40.3} \textcolor{red}{{\scriptsize(+2.1)}} &  \textbf{28.5}  \textcolor{red}{{\scriptsize(+5.0)}} \\ 
    \bottomrule[1.2pt]
    \end{tabular}
  \end{adjustbox}
\end{minipage}
\vspace{-15pt}
\end{table*}

\subsection{Ablation Study}


We conducted ablation studies to evaluate ReVMap's core components. Unless otherwise noted, these experiments used the onboard camera-only setting, with models trained on the nuScenes training new split and evaluated on the new split validation set.

\noindent\textbf{Contributions of Main Components.}

\textcolor{black}{Table~\ref{nusc_main_ablation} ablates the main components of ReVMap. The baseline uses a hand-crafted endpoint-distance rule for local-to-global association instead of learned connectivity estimation.}
Connectivity Estimation improves mGAP by 3.5\% and miECM by 4.8\%, demonstrating more reliable association than rule-based matching. Adding the Global Map Prior further improves mAP, mGAP, and miECM by 5.8\%, 3.9\%, and 5.1\%, showing the importance of accumulated map context. Uncertainty Estimation and Update further improve mGAP by 0.6\% and 1.1\%, strengthening reliability-aware prediction and global map fusion.

\noindent\textbf{Dependence on GT Boxes for Dynamic-object Removal.}
In the offboard camera--LiDAR setting, dynamic objects are removed from the aggregated LiDAR sweeps using GT 3D boxes.
Since public nuScenes detectors are trained on scenes that overlap the new split validation set, Table~\ref{tab:box_robust} instead perturbs the GT boxes with missed detections and box noise so that their class-wise AP matches that of MambaFusion~\cite{mambafusion}, and also evaluates a variant that removes no dynamic points.
mAP and mGAP drop by only 0.4\% and 0.5\% with perturbed boxes and by 0.7\% and 0.9\% without removal, showing that ReVMap does not critically depend on GT boxes.



\noindent\textbf{Effects of Uncertainty Update.}
Table~\ref{ablation_uncertainty} evaluates uncertainty-based Filtering and uncertainty-weighted Interpolation in the Uncertainty Update.
Filtering alone improves mGAP by 0.7\%, while Interpolation alone yields a 0.5\% gain, and combining both operations raises mGAP to 28.5\%, demonstrating their complementary roles in reliable global map update.


\begin{figure}[t] 
    \centering
    \includegraphics[width=0.75\linewidth]{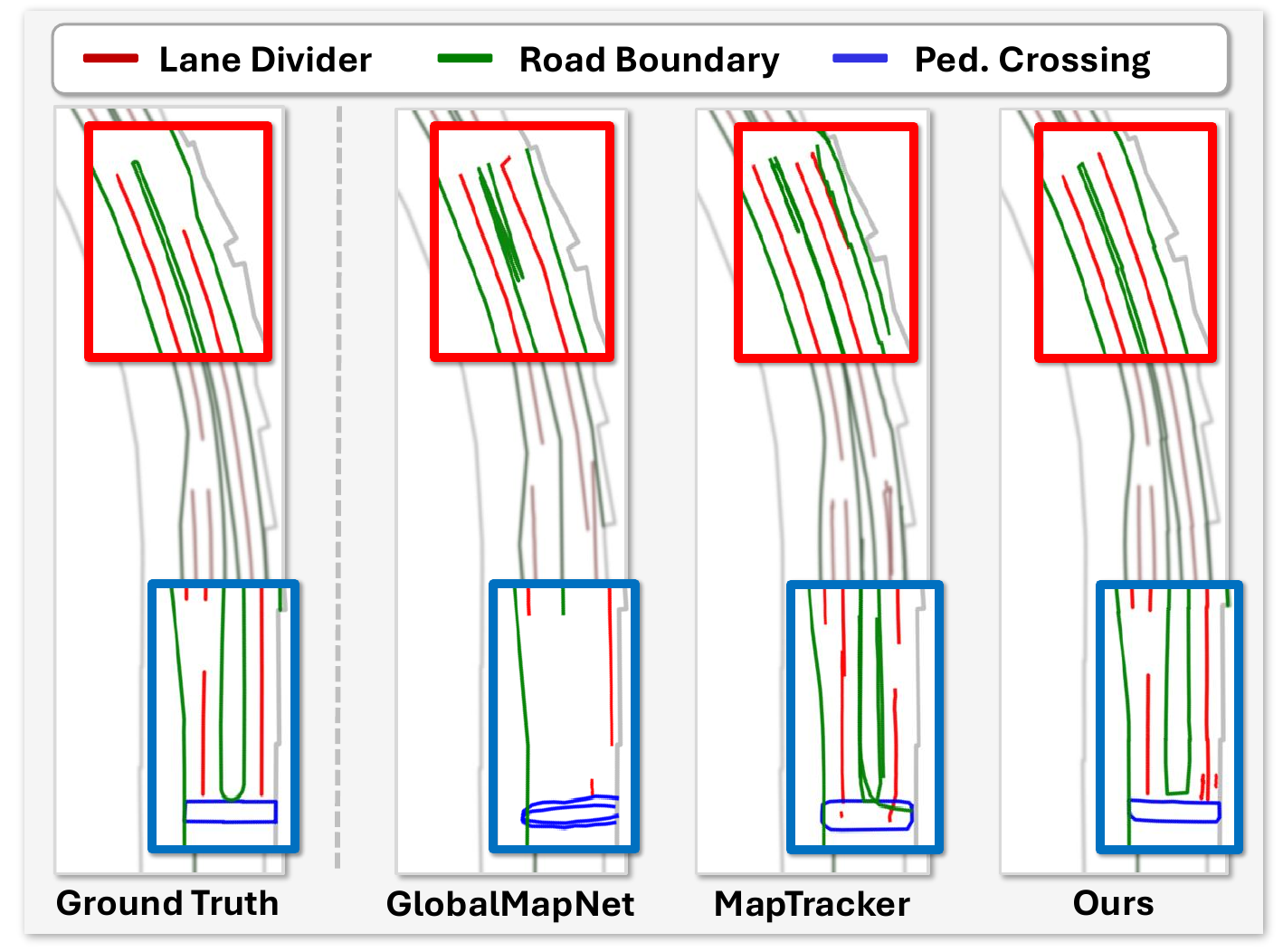}
    \captionof{figure}{\textbf{Qualitative results on the nuScenes new split validation set.}}
\label{fig:quali_c}
\end{figure}

\noindent\textbf{Plug-in Analysis of Global Construction.}
Table~\ref{tab:cross_analysis} analyzes our global map construction across different local predictors and existing global methods.
GlobalMapNet and MapTracker use their own local predictors trained without an initial map, and the same ReVMap local predictions are processed by GlobalMapNet, VMA, and our method, so only the global construction method changes within each group.
Combining our global construction with GlobalMapNet~\cite{globalmapnet} and MapTracker~\cite{maptracker} improves both mGAP and miECM, while replacing ours with GlobalMapNet or VMA~\cite{vma} under the same local predictor degrades mGAP by 5.0\% and 2.8\%, respectively.
This confirms that the gains generalize across local predictors and stem from the proposed connectivity estimation and uncertainty-aware update.

\begin{figure}[t] 
    \centering
    \includegraphics[width=\linewidth]{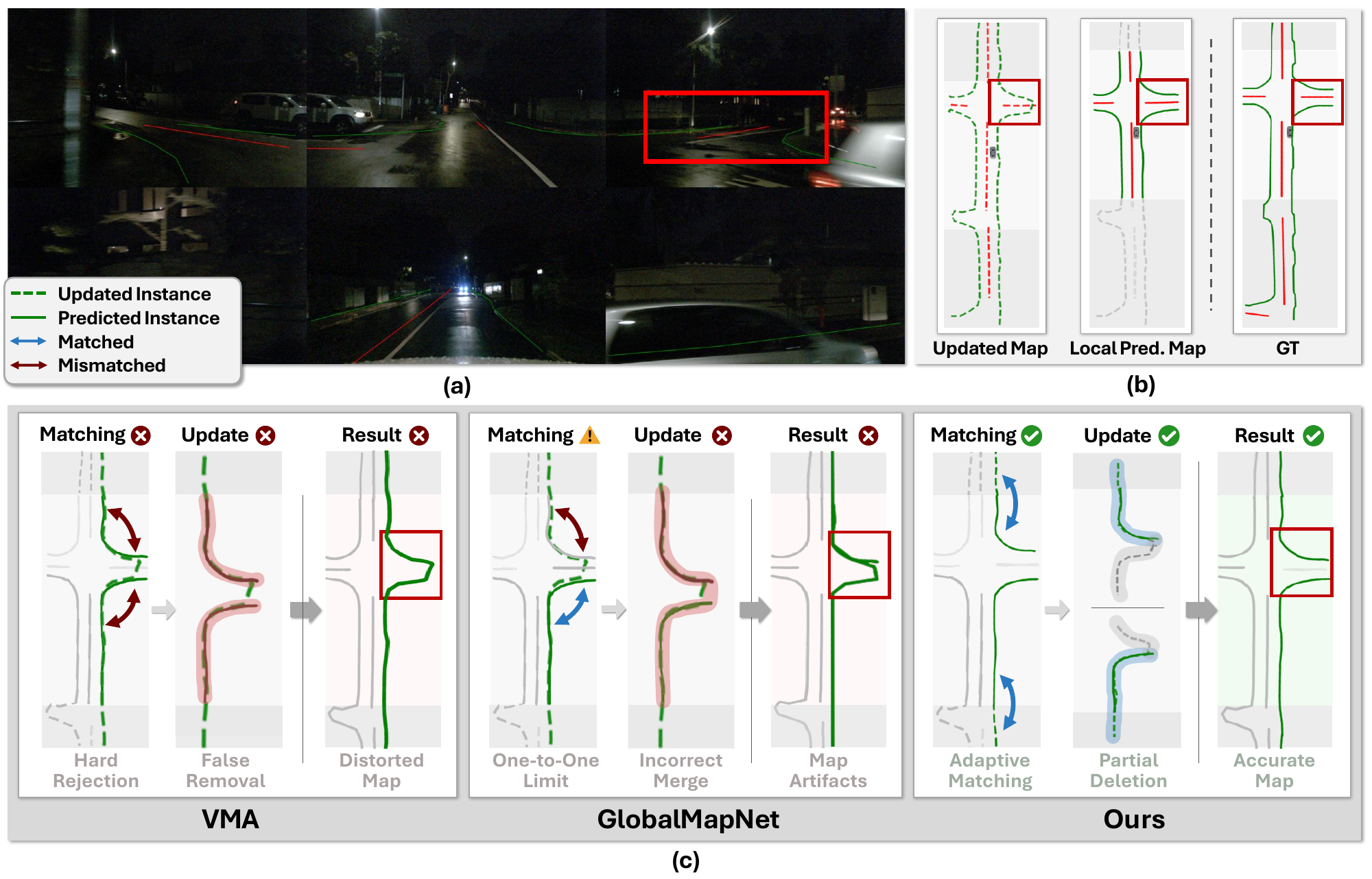}
    \vspace{-15pt}
    \caption{\textbf{Visualization of the Matching and Update Process.} 
(a) Multi-view camera inputs. 
(b) The accumulated global map, the current local prediction, and the ground-truth global map.
(c) Comparison of update processes between existing methods and the proposed ReVMap. 
}
\label{fig:quali_compare}
\end{figure}

\subsection{Qualitative Analysis}

\noindent\textbf{Qualitative Results Compared with SOTA Methods.}

Figure~\ref{fig:quali_c} compares ReVMap with GlobalMapNet~\cite{globalmapnet} and MapTracker~\cite{maptracker}. While previous methods exhibit topological inconsistencies, ReVMap produces more coherent and structurally consistent maps.
Additional qualitative results are provided in the Appendix.

\noindent\textbf{Qualitative Analysis of the Update Process.}
Figure~\ref{fig:quali_compare} illustrates the matching and update process of existing methods and the proposed method. VMA~\cite{vma} discards valid local instances and retains erroneous global geometry because its rule-based matching rejects pairs that fail endpoint-distance criteria. 
GlobalMapNet~\cite{globalmapnet} produces incorrect updates because it simply replaces only the overlapping regions between matched local and global instances. In contrast, ReVMap identifies revision targets through connectivity estimation and applies uncertainty-aware fusion, correcting erroneous segments without disrupting valid global structures.



\section{Discussion and Conclusion}
\noindent\textbf{Limitations and Future Work.}
\label{sec:limitations}
\textcolor{black}{Despite reducing manual effort in map auto-labeling, ReVMap remains supervised and relies on ground-truth vectorized maps to train its local map prediction and connectivity estimation modules. 
\textcolor{black}{In addition, the map update process relies on empirically selected thresholds for uncertainty filtering and geometric revision, which may require tuning for different datasets.}
Future work could reduce reliance on ground-truth maps by generating pseudo-labels from unlabeled sensor data, potentially leveraging neural reconstruction methods such as Gaussian splatting to provide geometric cues.}

\noindent\textbf{Conclusion.}
We presented ReVMap, a local-to-global framework for reliable vectorized global map construction. By combining uncertainty-guided prior reuse with connectivity-guided global map revision, ReVMap enables the correction of accumulated map errors while limiting their propagation into subsequent local predictions. Experiments on nuScenes and Argoverse2 demonstrate state-of-the-art performance in both local map prediction and global map construction, highlighting the effectiveness of reliability-aware map evolution.

\newpage


\bibliographystyle{plainnat}
\bibliography{references}

@article{maptrv2,
  title={Maptrv2: An end-to-end framework for online vectorized hd map construction},
  author={Liao, Bencheng and Chen, Shaoyu and Zhang, Yunchi and Jiang, Bo and Zhang, Qian and Liu, Wenyu and Huang, Chang and Wang, Xinggang},
  journal={International Journal of Computer Vision},
  volume={133},
  number={3},
  pages={1352--1374},
  year={2025},
  publisher={Springer}
}

@article{maptr,
  title={Maptr: Structured modeling and learning for online vectorized hd map construction},
  author={Liao, Bencheng and Chen, Shaoyu and Wang, Xinggang and Cheng, Tianheng and Zhang, Qian and Liu, Wenyu and Huang, Chang},
  journal={arXiv preprint arXiv:2208.14437},
  year={2022}
}

@inproceedings{mask2map,
  title={Mask2map: Vectorized hd map construction using bird’s eye view segmentation masks},
  author={Choi, Sehwan and Kim, Jungho and Shin, Hongjae and Choi, Jun Won},
  booktitle={European Conference on Computer Vision},
  pages={19--36},
  year={2024},
  organization={Springer}
}

@inproceedings{streammapnet,
  title={Streammapnet: Streaming mapping network for vectorized online hd map construction},
  author={Yuan, Tianyuan and Liu, Yicheng and Wang, Yue and Wang, Yilun and Zhao, Hang},
  booktitle={Proceedings of the IEEE/CVF Winter Conference on Applications of Computer Vision},
  pages={7356--7365},
  year={2024}
}

@inproceedings{uigenmap,
  title={Uncertainty-instructed structure injection for generalizable hd map construction},
  author={Liu, Xiaolu and Yang, Ruizi and Wang, Song and Li, Wentong and Chen, Junbo and Zhu, Jianke},
  booktitle={Proceedings of the Computer Vision and Pattern Recognition Conference},
  pages={22359--22368},
  year={2025}
}

@inproceedings{hrmapnet,
  title={Enhancing vectorized map perception with historical rasterized maps},
  author={Zhang, Xiaoyu and Liu, Guangwei and Liu, Zihao and Xu, Ningyi and Liu, Yunhui and Zhao, Ji},
  booktitle={European Conference on Computer Vision},
  pages={422--439},
  year={2024},
  organization={Springer}
}

@article{globalmapnet,
  title={Globalmapnet: An online framework for vectorized global hd map construction},
  author={Shi, Anqi and Cai, Yuze and Chen, Xiangyu and Pu, Jian and Fu, Zeyu and Lu, Hong},
  journal={arXiv preprint arXiv:2409.10063},
  year={2024}
}

@inproceedings{maptracker,
  title={Maptracker: Tracking with strided memory fusion for consistent vector hd mapping},
  author={Chen, Jiacheng and Wu, Yuefan and Tan, Jiaqi and Ma, Hang and Furukawa, Yasutaka},
  booktitle={European Conference on Computer Vision},
  pages={90--107},
  year={2024},
  organization={Springer}
}

@article{vma,
  title={Vma: Divide-and-conquer vectorized map annotation system for large-scale driving scene},
  author={Chen, Shaoyu and Zhang, Yunchi and Liao, Bencheng and Xie, Jiafeng and Cheng, Tianheng and Sui, Wei and Zhang, Qian and Huang, Chang and Liu, Wenyu and Wang, Xinggang},
  journal={arXiv preprint arXiv:2304.09807},
  year={2023}
}

@article{egcvmap,
  title={End-to-End Generation of City-Scale Vectorized Maps by Crowdsourced Vehicles},
  author={Feng, Zebang and Fan, Miao and Liu, Bao and Xu, Shengtong and Xiong, Haoyi},
  journal={arXiv preprint arXiv:2507.08901},
  year={2025}
}

@article{safedrive,
  title={SafeDrive: Fine-Grained Safety Reasoning for End-to-End Driving in a Sparse World},
  author={Kim, Jungho and Oh, Jiyong and Yu, Seunghoon and Shin, Hongjae and Kwak, Donghyuk and Choi, Jun Won},
  journal={arXiv preprint arXiv:2602.18887},
  year={2026}
}

@inproceedings{nuscenes,
  title={nuscenes: A multimodal dataset for autonomous driving},
  author={Caesar, Holger and Bankiti, Varun and Lang, Alex H and Vora, Sourabh and Liong, Venice Erin and Xu, Qiang and Krishnan, Anush and Pan, Yu and Baldan, Giancarlo and Beijbom, Oscar},
  booktitle={Proceedings of the IEEE/CVF conference on computer vision and pattern recognition},
  pages={11621--11631},
  year={2020}
}

@inproceedings{argoverse2,
  title={Argoverse 2: Next Generation Datasets for Self-Driving Perception and Forecasting},
  author={Wilson, Benjamin and Qi, William and Agarwal, Tanmay and Lambert, John and Singh, Jagjeet and Khandelwal, Siddhesh and Pan, Bowen and Kumar, Ratnesh and Hartnett, Andrew and Pontes, Jhony Kaesemodel and others},
  booktitle={Thirty-fifth Conference on Neural Information Processing Systems Datasets and Benchmarks Track (Round 2)}
}

@article{camav2,
  title={Camav2: A vision-centric approach for static map element annotation},
  author={Chen, Shiyuan and Zhang, Jiaxin and Mei, Ruohong and Cai, Yingfeng and Yin, Haoran and Sui, Wei and Yang, Cong},
  journal={IEEE Transactions on Intelligent Transportation Systems},
  year={2026},
  publisher={IEEE}
}

@inproceedings{cama,
  title={A vision-centric approach for static map element annotation},
  author={Zhang, Jiaxin and Chen, Shiyuan and Yin, Haoran and Mei, Ruohong and Liu, Xuan and Yang, Cong and Zhang, Qian and Sui, Wei},
  booktitle={2024 IEEE International Conference on Robotics and Automation (ICRA)},
  pages={15861--15867},
  year={2024},
  organization={IEEE}
}

@inproceedings{thma,
  title={Thma: Tencent hd map ai system for creating hd map annotations},
  author={Tang, Kun and Cao, Xu and Cao, Zhipeng and Zhou, Tong and Li, Erlong and Liu, Ao and Zou, Shengtao and Liu, Chang and Mei, Shuqi and Sizikova, Elena and others},
  booktitle={Proceedings of the AAAI Conference on Artificial Intelligence},
  volume={37},
  number={13},
  pages={15585--15593},
  year={2023}
}

@inproceedings{pivotnet,
  title={Pivotnet: Vectorized pivot learning for end-to-end hd map construction},
  author={Ding, Wenjie and Qiao, Limeng and Qiu, Xi and Zhang, Chi},
  booktitle={Proceedings of the IEEE/CVF International Conference on Computer Vision},
  pages={3672--3682},
  year={2023}
}

@inproceedings{bemapnet,
  title={End-to-end vectorized hd-map construction with piecewise bezier curve},
  author={Qiao, Limeng and Ding, Wenjie and Qiu, Xi and Zhang, Chi},
  booktitle={Proceedings of the IEEE/CVF Conference on Computer Vision and Pattern Recognition},
  pages={13218--13228},
  year={2023}
}

@inproceedings{mapqr,
  title={Leveraging enhanced queries of point sets for vectorized map construction},
  author={Liu, Zihao and Zhang, Xiaoyu and Liu, Guangwei and Zhao, Ji and Xu, Ningyi},
  booktitle={European Conference on Computer Vision},
  pages={461--477},
  year={2024},
  organization={Springer}
}

@inproceedings{dumapnet,
  title={Dumapnet: An end-to-end vectorization system for city-scale lane-level map generation},
  author={Xia, Deguo and Zhang, Weiming and Liu, Xiyan and Zhang, Wei and Gong, Chenting and Huang, Jizhou and Yang, Mengmeng and Yang, Diange},
  booktitle={Proceedings of the 30th ACM SIGKDD Conference on Knowledge Discovery and Data Mining},
  pages={6015--6024},
  year={2024}
}

@inproceedings{nmp,
  title={Neural map prior for autonomous driving},
  author={Xiong, Xuan and Liu, Yicheng and Yuan, Tianyuan and Wang, Yue and Wang, Yilun and Zhao, Hang},
  booktitle={Proceedings of the IEEE/CVF Conference on Computer Vision and Pattern Recognition},
  pages={17535--17544},
  year={2023}
}

@inproceedings{presight,
  title={Presight: Enhancing autonomous vehicle perception with city-scale nerf priors},
  author={Yuan, Tianyuan and Mao, Yucheng and Yang, Jiawei and Liu, Yicheng and Wang, Yue and Zhao, Hang},
  booktitle={European Conference on Computer Vision},
  pages={323--339},
  year={2024},
  organization={Springer}
}

@inproceedings{rpred,
  title={R-pred: Two-stage motion prediction via tube-query attention-based trajectory refinement},
  author={Choi, Sehwan and Kim, Jungho and Yun, Junyong and Choi, Jun Won},
  booktitle={Proceedings of the IEEE/CVF International Conference on Computer Vision},
  pages={8525--8535},
  year={2023}
}

@inproceedings{uniad,
  title={Planning-oriented autonomous driving},
  author={Hu, Yihan and Yang, Jiazhi and Chen, Li and Li, Keyu and Sima, Chonghao and Zhu, Xizhou and Chai, Siqi and Du, Senyao and Lin, Tianwei and Wang, Wenhai and others},
  booktitle={Proceedings of the IEEE/CVF conference on computer vision and pattern recognition},
  pages={17853--17862},
  year={2023}
}

@inproceedings{vad,
  title={Vad: Vectorized scene representation for efficient autonomous driving},
  author={Jiang, Bo and Chen, Shaoyu and Xu, Qing and Liao, Bencheng and Chen, Jiajie and Zhou, Helong and Zhang, Qian and Liu, Wenyu and Huang, Chang and Wang, Xinggang},
  booktitle={Proceedings of the IEEE/CVF International Conference on Computer Vision},
  pages={8340--8350},
  year={2023}
}

@inproceedings{mvmap,
  title={Mv-map: Offboard hd-map generation with multi-view consistency},
  author={Xie, Ziyang and Pang, Ziqi and Wang, Yu-Xiong},
  booktitle={Proceedings of the IEEE/CVF International Conference on Computer Vision},
  pages={8658--8668},
  year={2023}
}

@inproceedings{lss,
  title={Lift, splat, shoot: Encoding images from arbitrary camera rigs by implicitly unprojecting to 3d},
  author={Philion, Jonah and Fidler, Sanja},
  booktitle={European conference on computer vision},
  pages={194--210},
  year={2020},
  organization={Springer}
}

@inproceedings{mask2former,
  title={Masked-attention mask transformer for universal image segmentation},
  author={Cheng, Bowen and Misra, Ishan and Schwing, Alexander G and Kirillov, Alexander and Girdhar, Rohit},
  booktitle={Proceedings of the IEEE/CVF conference on computer vision and pattern recognition},
  pages={1290--1299},
  year={2022}
}

@inproceedings{mgmap,
  title={Mgmap: Mask-guided learning for online vectorized hd map construction},
  author={Liu, Xiaolu and Wang, Song and Li, Wentong and Yang, Ruizi and Chen, Junbo and Zhu, Jianke},
  booktitle={Proceedings of the IEEE/CVF Conference on Computer Vision and Pattern Recognition},
  pages={14812--14821},
  year={2024}
}

@inproceedings{vectormapnet,
  title={Vectormapnet: End-to-end vectorized hd map learning},
  author={Liu, Yicheng and Yuan, Tianyuan and Wang, Yue and Wang, Yilun and Zhao, Hang},
  booktitle={International conference on machine learning},
  pages={22352--22369},
  year={2023},
  organization={PMLR}
}

@inproceedings{hdmapnet,
  title={Hdmapnet: An online hd map construction and evaluation framework},
  author={Li, Qi and Wang, Yue and Wang, Yilun and Zhao, Hang},
  booktitle={2022 International Conference on Robotics and Automation (ICRA)},
  pages={4628--4634},
  year={2022},
  organization={IEEE}
}

@inproceedings{uncertaintyTrajectory,
  title={Producing and leveraging online map uncertainty in trajectory prediction},
  author={Gu, Xunjiang and Song, Guanyu and Gilitschenski, Igor and Pavone, Marco and Ivanovic, Boris},
  booktitle={Proceedings of the IEEE/CVF Conference on Computer Vision and Pattern Recognition},
  pages={14521--14530},
  year={2024}
}

@article{topoboundary,
  title={Topo-boundary: A benchmark dataset on topological road-boundary detection using aerial images for autonomous driving},
  author={Xu, Zhenhua and Sun, Yuxiang and Liu, Ming},
  journal={IEEE Robotics and Automation Letters},
  volume={6},
  number={4},
  pages={7248--7255},
  year={2021},
  publisher={IEEE}
}

@article{vit,
  title={An image is worth 16x16 words: Transformers for image recognition at scale},
  author={Dosovitskiy, Alexey and Beyer, Lucas and Kolesnikov, Alexander and Weissenborn, Dirk and Zhai, Xiaohua and Unterthiner, Thomas and Dehghani, Mostafa and Minderer, Matthias and Heigold, Georg and Gelly, Sylvain and others},
  journal={arXiv preprint arXiv:2010.11929},
  year={2020}
}

@article{second,
  title={Second: Sparsely embedded convolutional detection},
  author={Yan, Yan and Mao, Yuxing and Li, Bo},
  journal={Sensors},
  volume={18},
  number={10},
  pages={3337},
  year={2018},
  publisher={MDPI}
}

@article{uniprev,
  title={Uni-PrevPredMap: Extending PrevPredMap to a unified framework of prior-informed modeling for online vectorized HD map construction},
  author={Peng, Nan and Zhou, Xun and Wang, Mingming and Chen, Guisong and Xu, Wenqi},
  journal={arXiv preprint arXiv:2504.06647},
  year={2025}
}

@inproceedings{unimapgen,
  title={Unimapgen: A generative framework for large-scale map construction from multi-modal data},
  author={Yuan, Yujian and Wu, Changjie and Chang, Xinyuan and Wang, Sijin and Zhang, Hang and Liang, Shiyi and Zeng, Shuang and Xu, Mu},
  booktitle={Proceedings of the AAAI Conference on Artificial Intelligence},
  volume={40},
  number={15},
  pages={12277--12285},
  year={2026}
}

@inproceedings{icmapper,
  title={IC-Mapper: Instance-Centric Spatio-Temporal Modeling for Online Vectorized Map Construction},
  author={Zhu, Jiangtong and Yang, Zhao and Shi, Yinan and Fang, Jianwu and Xue, Jianru},
  booktitle={Proceedings of the 32nd ACM International Conference on Multimedia},
  pages={9961--9969},
  year={2024}
}

@inproceedings{focal,
  title={Focal loss for dense object detection},
  author={Lin, Tsung-Yi and Goyal, Priya and Girshick, Ross and He, Kaiming and Doll{\'a}r, Piotr},
  booktitle={Proceedings of the IEEE international conference on computer vision},
  pages={2980--2988},
  year={2017}
}

@inproceedings{bevfusion,
  title={Bevfusion: Multi-task multi-sensor fusion with unified bird's-eye view representation},
  author={Liu, Zhijian and Tang, Haotian and Amini, Alexander and Yang, Xinyu and Mao, Huizi and Rus, Daniela L and Han, Song},
  booktitle={2023 IEEE international conference on robotics and automation (ICRA)},
  pages={2774--2781},
  year={2023},
  organization={IEEE}
}

@inproceedings{mapillary,
  title={The mapillary vistas dataset for semantic understanding of street scenes},
  author={Neuhold, Gerhard and Ollmann, Tobias and Rota Bulo, Samuel and Kontschieder, Peter},
  booktitle={Proceedings of the IEEE international conference on computer vision},
  pages={4990--4999},
  year={2017}
}

@inproceedings{hivt,
  title={Hivt: Hierarchical vector transformer for multi-agent motion prediction},
  author={Zhou, Zikang and Ye, Luyao and Wang, Jianping and Wu, Kui and Lu, Kejie},
  booktitle={Proceedings of the IEEE/CVF conference on computer vision and pattern recognition},
  pages={8823--8833},
  year={2022}
}

@article{smart,
  title={Smart: Scalable multi-agent real-time motion generation via next-token prediction},
  author={Wu, Wei and Feng, Xiaoxin and Gao, Ziyan and Kan, Yuheng},
  journal={Advances in Neural Information Processing Systems},
  volume={37},
  pages={114048--114071},
  year={2024}
}

@article{openlanev2,
  title={Openlane-v2: A topology reasoning benchmark for unified 3d hd mapping},
  author={Wang, Huijie and Li, Tianyu and Li, Yang and Chen, Li and Sima, Chonghao and Liu, Zhenbo and Wang, Bangjun and Jia, Peijin and Wang, Yuting and Jiang, Shengyin and others},
  journal={Advances in Neural Information Processing Systems},
  volume={36},
  pages={18873--18884},
  year={2023}
}

@inproceedings{lanegcn,
  title={Learning lane graph representations for motion forecasting},
  author={Liang, Ming and Yang, Bin and Hu, Rui and Chen, Yun and Liao, Renjie and Feng, Song and Urtasun, Raquel},
  booktitle={European Conference on Computer Vision},
  pages={541--556},
  year={2020},
  organization={Springer}
}

@inproceedings{incrementalfusing,
  title={An incremental fusing method for high-definition map updating},
  author={Liu, Yinqi and Song, Mengxuan and Guo, Yafeng},
  booktitle={2019 IEEE International Conference on Systems, Man and Cybernetics (SMC)},
  pages={4251--4256},
  year={2019},
  organization={IEEE}
}

@inproceedings{mambafusion,
  title={{MambaFusion}: Height-Fidelity Dense Global Fusion for Multi-modal {3D} Object Detection},
  author={Wang, Hanshi and Gao, Jin and Hu, Weiming and Zhang, Zhipeng},
  booktitle={Proceedings of the IEEE/CVF International Conference on Computer Vision},
  pages={26664--26674},
  year={2025}
}

@article{instagram,
  title={Instagram: Instance-level graph modeling for vectorized hd map learning},
  author={Shin, Juyeb and Jeong, Hyeonjun and Rameau, Francois and Kum, Dongsuk},
  journal={IEEE Transactions on Intelligent Transportation Systems},
  volume={26},
  number={2},
  pages={1889--1899},
  year={2025},
  publisher={IEEE}
}

\newpage


\begin{center}
    {\LARGE \bfseries Supplementary Materials for ReVMap}
\end{center}

\vspace{1em}

\section{Additional Details for ReVMap}
\subsection{Evaluation Metrics}
\noindent\textbf{Chamfer-based AP Metrics.}
We evaluate mAP, mGAP, and multi-trip mGAP using the same Chamfer-distance-based AP computation. 
AP is computed at distance thresholds of 0.5, 1.0, and 1.5 meters for lane divider, pedestrian crossing, and road boundary classes. 
The final score is obtained by averaging AP over thresholds and classes.

\noindent\textbf{Instance-level ECM.}
ECM~\cite{topoboundary} is designed to penalize cases where a single GT element is covered by multiple fragmented predictions. 
To compute this, predicted map elements are rasterized, and each prediction mask is assigned to the GT mask closest to most of its pixels. 
It then scores each GT mask using the coverage and entropy of its assigned prediction masks.
However, overlapping fragmented predictions are merged into a single component after rasterization, causing ECM to treat multiple fragments as a single prediction and thereby under-penalize fragmented predictions.
We therefore rasterize each predicted polyline separately, keeping fragments as individual prediction masks, and compute ECM in the same manner. 
For each GT mask $M_j$, $C_j$ measures the entropy of the normalized sizes of its assigned prediction masks, while $\alpha_j$ measures their coverage of $M_j$. 
The per-class instance-level ECM is
\[
\mathrm{iECM}_c
=
\sum_j
\frac{|M_j|}{\sum_{j'}|M_{j'}|}
\alpha_j e^{-C_j}.
\]
We report $\mathrm{miECM}$ by averaging over classes, for both scene-level and multi-trip global maps.


\noindent\textbf{TOP Metric.}
To evaluate the local-to-global connectivity prediction performance in the Connectivity Estimation Module, we adapt the TOP metric~\cite{openlanev2}, proposed for lane topology reasoning.
The original TOP metric uses Fréchet-distance-based matching to align predicted instances with ground-truth instances before computing per-instance mean average precision. 
We instead evaluate the binary connectivity matrix between local predictions and global instances using Chamfer-distance-based matching with a distance threshold of 1.5\,m. 
The connectivity ground truth is defined based on matched ground-truth instances in each region, assigning connectivity when a local instance and a global instance are matched to the same ground-truth instance. It is defined over all four directional combinations between the original and reverse directions of local and global instances, with only the convergent matching pair labeled as positive.

\subsection{Implementation Details} 
\label{sec:Implementation}
For nuScenes, input images are resized by 0.5. For Argoverse2, seven camera images are padded to $2048 \times 2048$ and resized by 0.3.
For image-to-BEV transformation, we use an LSS-based image-to-BEV module with a BEV grid resolution of 0.3 m. The local perception range is set to $[-30, 30, -15, 15]$m, and global instances for connectivity estimation are retrieved within $[-45, 45, -30, 30]$m, following the longitudinal and lateral order. Mask2Former~\cite{mask2former} is pretrained on Mapillary Vistas V2~\cite{mapillary} for dynamic-region masking.
ULG uses 100 instance queries, each with 20 point queries, so each local map instance contains 20 points. Global instances in the Global Prior Encoder and Connectivity Module are also resampled to 20 points. The Global Prior Encoder and Local Map Decoder consist of 2 and 6 layers, respectively. The Connectivity Module consists of an encoder with a single layer and a decoder with 6 layers. For uncertainty filtering, the uncertainty thresholds for lane divider, pedestrian crossing, and road boundary are empirically set to 1.6, 1.4, and 2.8, respectively. Before fusing a matched local-global instance pair, we trim the matched global instance by discarding segments farther than 1.5 m from the local instance. We use AdamW with an initial learning rate of $6 \times 10^{-4}$ and weight decay of 0.01, cosine learning-rate decay with 500 warm-up iterations, and four A6000 GPUs with a total batch size of 16.

\newpage 

\section{Supplementary Experimental Results}

\begin{table*}[h!] 
\centering
\begin{minipage}{0.45\textwidth}
    \centering
    \captionof{table}{Impact of ground-truth preprocessing on local map construction. All models are evaluated under the camera-only onboard setting.}
\begin{adjustbox}{width=\textwidth}
\begin{tabular}{l|ccc|c}
\toprule[1.2pt]
Method & $div$ & $ped$ & $bou$ & mAP \\
\midrule[0.7pt]
MapTRv2 (merged GT) & 26.5 & 37.2 & 46.1 & 36.6 \\
MapTRv2 (split GT) & 29.3 & 42.0 & 45.7 & 39.0 \\
ReVMap (Ours) & 33.7 & 49.6 & 53.4 & 45.6 \\
\bottomrule[1.2pt]
\end{tabular}
\end{adjustbox}
    \label{tab:nusc_maptrv2}
\end{minipage}
\hfill
\begin{minipage}{0.48\textwidth}
    \centering
    \captionof{table}{Ablation study for evaluating the effect of the Connectivity Estimation submodules. ``Encoder'' denotes the bidirectional Map Instance Tokenizer and ``Decoder'' refers to the Graph-based Connectivity Predictor.}
\begin{adjustbox}{width=\textwidth}
    \begin{tabular}{cc|ccc|c}
    \toprule[1.2pt]
     & & \multicolumn{4}{c}{TOP Metrics} \\
    \cmidrule(lr){3-6}
    Encoder & Decoder & $div$ & $ped$ & $bou$ & $mean$ \\
    \midrule[0.7pt]
     &  & 58.2 & 58.3 & 56.9 & 57.8 \\ 
    \checkmark &  & 61.0 & 62.5 & 57.1 & 60.2 \\ 
     & \checkmark & 61.8 & 63.4 & 57.0 & 60.7 \\ 
    \checkmark & \checkmark & \textbf{62.7} & \textbf{65.8} & \textbf{57.4} & \textbf{61.9} \\ 
    \bottomrule[1.2pt]
    \end{tabular}
\end{adjustbox}
    \label{tab:nusc_connectivity}
\end{minipage}
\end{table*}

\subsection{Baseline Analysis with Consistent Pedestrian Crossing Ground Truth}
Recent local map construction methods use different ground-truth definitions for pedestrian crossings. Several methods~\cite{mapqr, maptr, maptrv2, mask2map} merge overlapping pedestrian crossing polygon fragments into a single polygon and resample it into a fixed number of points. In contrast, StreamMapNet~\cite{streammapnet} and MapTracker~\cite{maptracker} use a different preprocessing strategy that avoids indiscriminate merging of pedestrian crossing fragments and produces more consistent pedestrian crossing instances.

Following this StreamMapNet-style preprocessing, we conduct an ablation study to clarify whether the performance gain comes from ground-truth preprocessing or from the proposed architecture. 
As shown in Table~\ref{tab:nusc_maptrv2}, changing the MapTRv2 baseline from merged GT to split GT notably improves pedestrian crossing AP from 37.2\% to 42.0\%, showing that the pedestrian crossing annotation definition has a significant impact on this category. Under the same split-GT setting, our model further improves the strengthened baseline and achieves the best performance across all categories, increasing mAP from 39.0\% to 45.6\%.



\subsection{Effect of the Connectivity Estimation submodules.}

Table~\ref{tab:nusc_connectivity} evaluates the two submodules of the Connectivity Estimation. 
In the baseline configuration, both submodules are replaced with simple MLPs. 
Adding either submodule alone improves the TOP metric by 2.4\% and 2.9\%, respectively, and applying both together yields a further gain to 61.9\%, showing that the two submodules are complementary. 
The bidirectional encoder captures the point ordering direction at each end of a polyline, while the graph-based decoder models geometric proximity between candidate pairs.

\section{Extensive Qualitative Results}
\begin{figure}[!h] 
    \centering
    \includegraphics[width=0.95\linewidth]{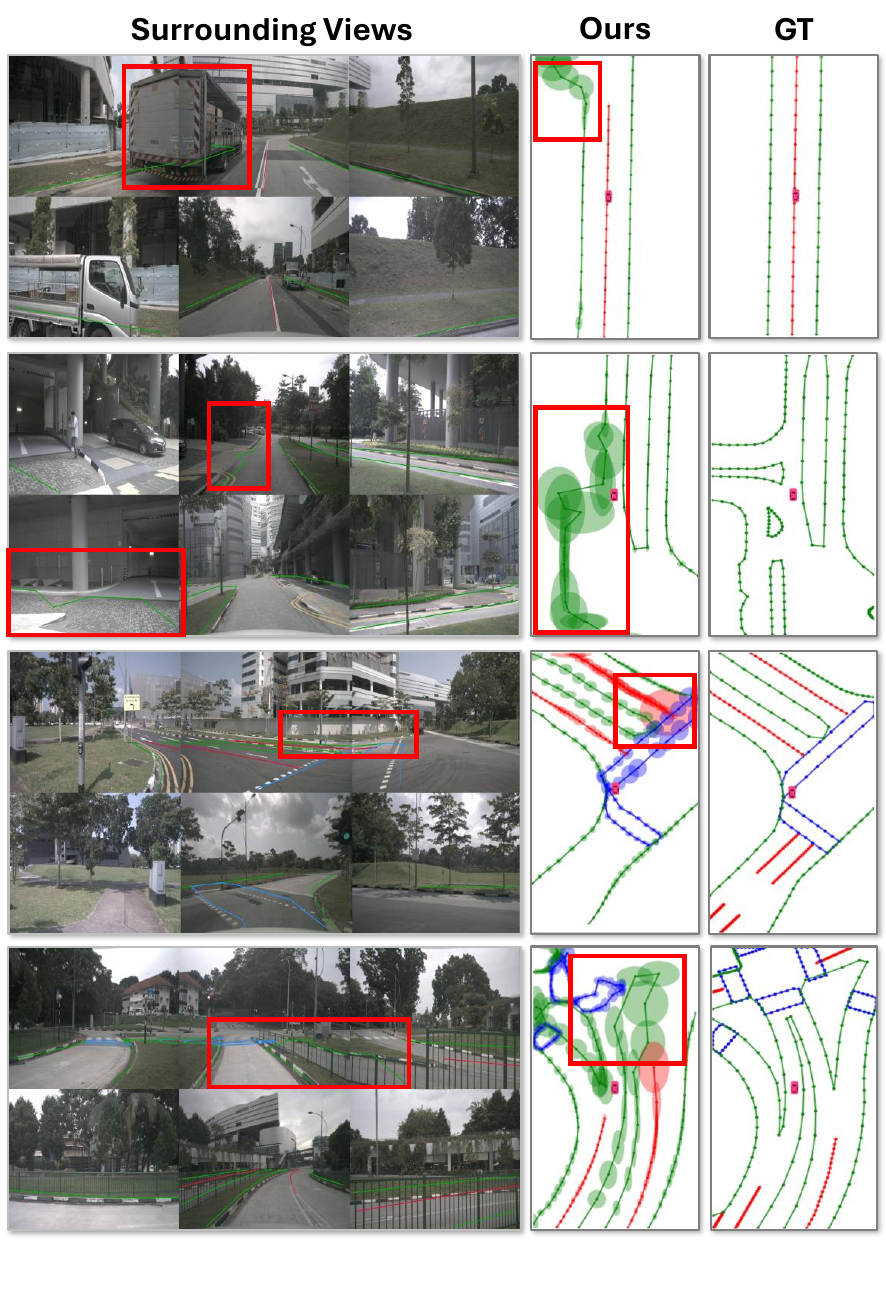}
    \caption{\textbf{Qualitative visualization of uncertainty-aware local vectorized maps on the nuScenes new split validation set.} 
    }
    \label{fig:unc_vis}
\end{figure}

\begin{figure}[!h] 
    \centering
    \includegraphics[width=\linewidth]{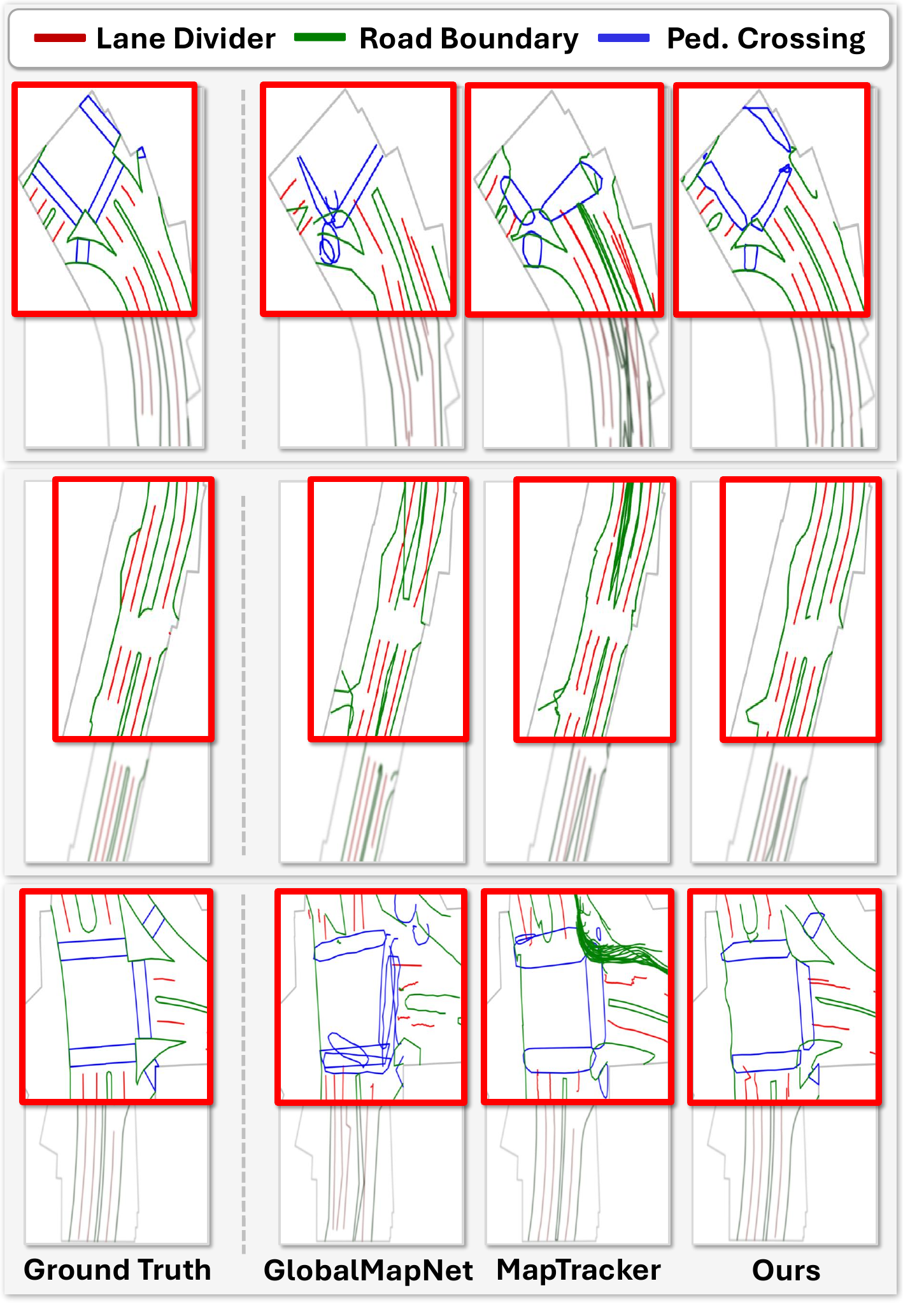}
    \caption{\textbf{Additional qualitative results on the nuScenes new split validation set.} 
    }
    \label{fig:quali_appendix}
\end{figure}

\begin{figure}[!h] 
    \centering
    \includegraphics[width=\linewidth]{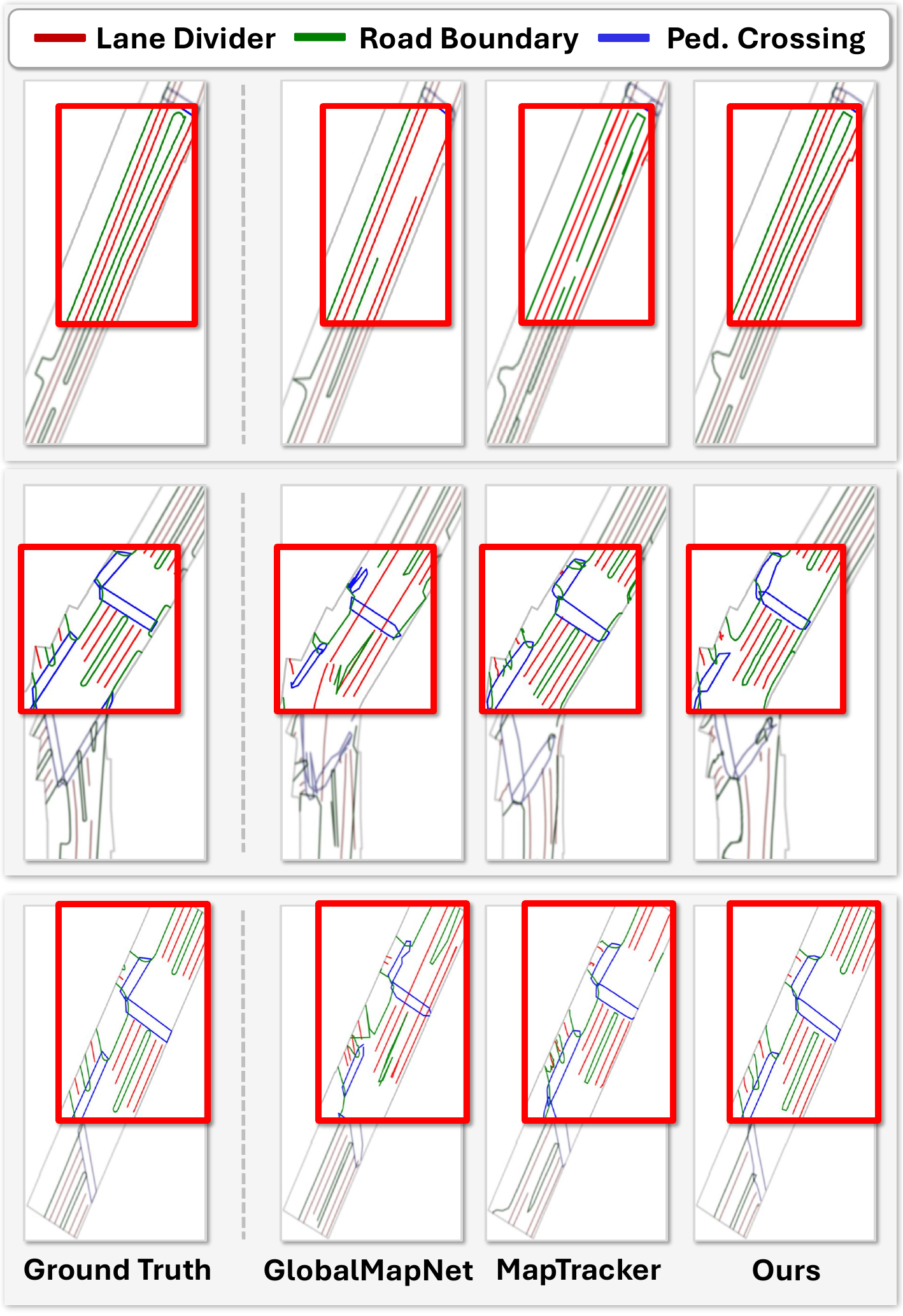}
    \caption{\textbf{Qualitative results on the Argoverse2 new split validation set.} 
    }
    \label{fig:quali_appendix_av2}
\end{figure}

\begin{figure}[!h] 
    \centering
    \includegraphics[width=1.0\linewidth]{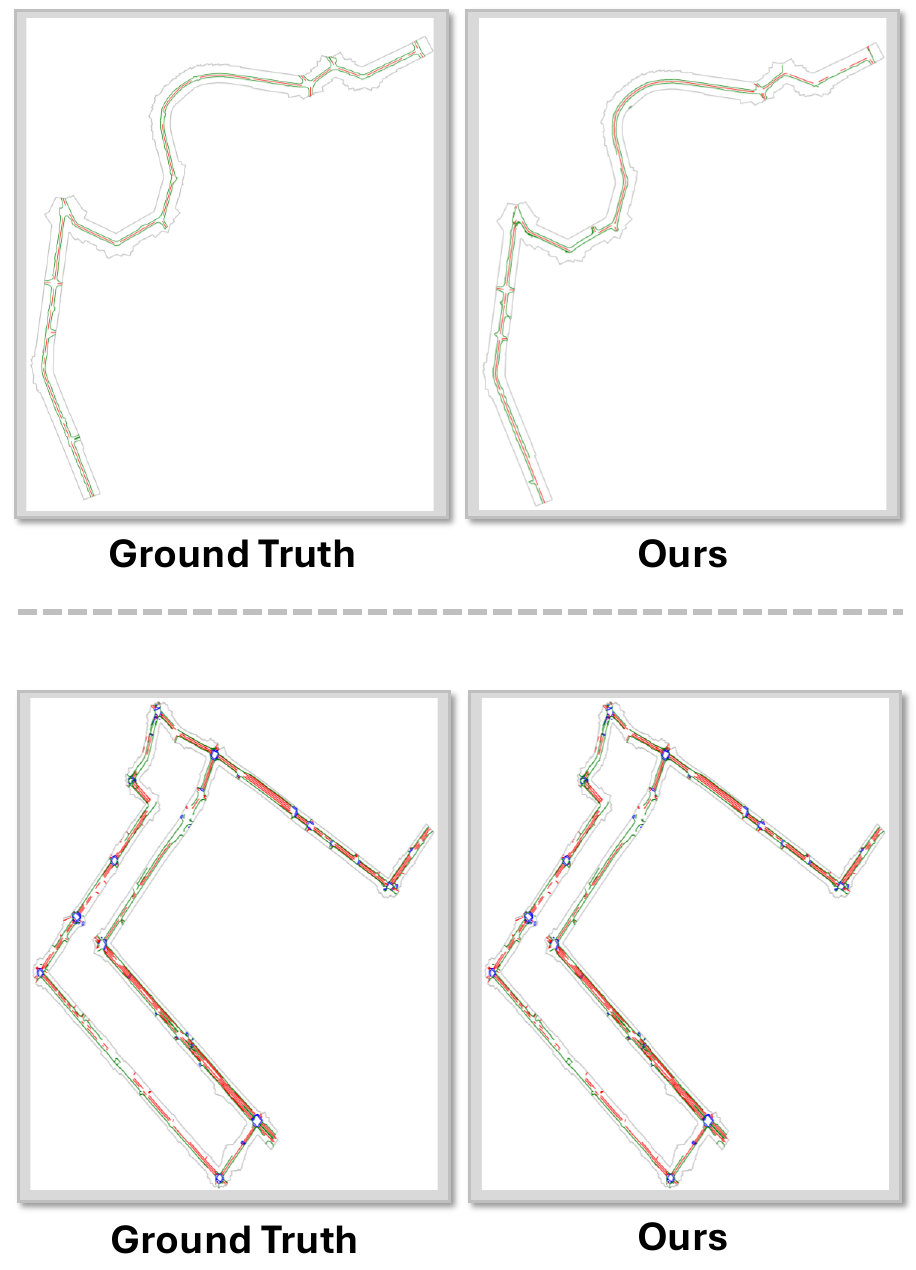}
    \caption{\textbf{Qualitative results of multi-trip vectorized global map construction on the nuScenes new split validation set.} 
    }
    \label{fig:globalscale_1}
\end{figure}

\begin{figure}[!h] 
    \centering
    \includegraphics[width=1.0\linewidth]{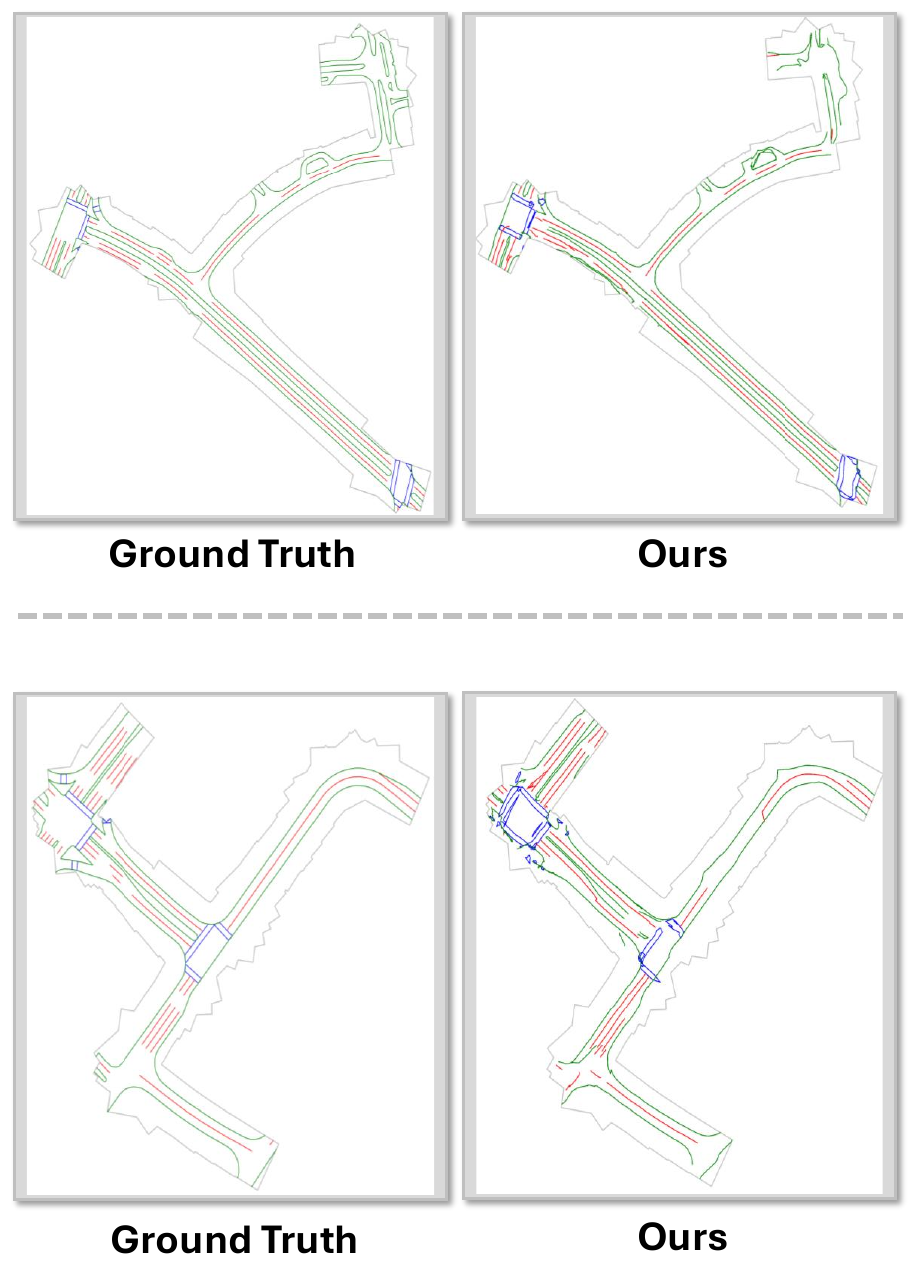}
    \caption{\textbf{Qualitative results of multi-trip vectorized global map construction on the nuScenes new split validation set.} 
    }
    \label{fig:globalscale_2}
\end{figure}

Figure~\ref{fig:unc_vis} visualizes the predicted local HD map together with its point-wise uncertainty on the nuScenes new split validation set. The uncertainty estimates concentrate on visually ambiguous and occluded regions, showing that ULG captures localization reliability.
Figures~\ref{fig:quali_appendix} and~\ref{fig:quali_appendix_av2} present additional qualitative results on the nuScenes and Argoverse2 new split validation sets, respectively, showing the consistency of the proposed method across scene layouts. Figures~\ref{fig:globalscale_1} and~\ref{fig:globalscale_2} provide comparisons between the reconstructed multi-trip global maps and the corresponding ground-truth maps, demonstrating that the proposed method produces globally coherent map structures over large-scale driving areas. 

\clearpage












\end{document}